%% file: main.tex
\documentclass[10pt,twocolumn,letterpaper]{article}
\usepackage{graphicx}
\usepackage{wacv}                    % CAMERA-READY version
\usepackage{algorithm}
\usepackage{algpseudocode}
\usepackage{xcolor}
\usepackage{caption}
\usepackage{array}
\usepackage{makecell}
\usepackage{multirow}
\usepackage{amssymb}
\usepackage{adjustbox}
\usepackage{stfloats} % Allow full-width figures at the bottom of a two-column page.

\definecolor{cvprblue}{rgb}{0.21,0.49,0.74}
\definecolor{raspberry}{rgb}{0.89, 0.04, 0.36} 

\algrenewcommand\algorithmicrequire{\textbf{Input:}}
\algrenewcommand\algorithmicensure{\textbf{Output:}}
\algrenewcommand{\algorithmiccomment}[1]{\hfill\textcolor{gray}{\textit{\# #1}}}


\definecolor{wacvblue}{rgb}{0.21,0.49,0.74}
\usepackage[pagebackref,breaklinks,colorlinks,allcolors=wacvblue]{hyperref}

\def\wacvPaperID{1936} % *** Enter the WACV Paper ID here
\def\confName{WACV}
\def\confYear{2027}

\title{Ray-SCR: Generalizing Scene Coordinate Regression Across Camera Models}

\author{Jaemin Cho\textsuperscript{1}\qquad
Oleksii Nasypanyi\textsuperscript{1}\qquad
Francois Rameau\textsuperscript{2}\\
\textsuperscript{1}Stony Brook University\qquad
\textsuperscript{2}SUNY Korea\\
{\tt\small \{jaemin.cho,oleksii.nasypanyi\}@stonybrook.edu}\quad
{\tt\small francois.rameau@sunykorea.ac.kr}
}

\begin{document}
 \maketitle
 \input{Section/0_Abstract}    
 \input{Section/1_Introduction}
 \input{Section/2_Related_Work}
 \input{Section/3_Methods_updated}
 \input{Section/4_Experiments}
 \input{Section/5_Conclusion}

\section*{Acknowledgements}
{
This work was supported by the National Research Foundation of Korea (RS-2024-0035-8054) and by IITP-2026-RS-2023-00259678.\par
}

\clearpage

{
    \small
    \bibliographystyle{ieeenat_fullname}
    \bibliography{main}
}

\clearpage
\setcounter{page}{1}
\maketitlesupplementary
\appendix
\input{Appendix}

\end{document}

%% file: Section/0_Abstract.tex
\begin{abstract}

Among visual localization strategies, scene coordinate regression (SCR) has emerged as an attractive approach for its compactness, accuracy, and speed. Despite these advantages, SCR models generalize poorly across substantially different camera models, such as omnidirectional or strongly rectified imagery.  This limitation arises because, changes in field of view, lens distortion, or rectification-induced resampling can make the same scene content appear differently depending on its pixel position, degrading localization accuracy. To cope with this limitation, we introduce Ray-SCR, a lightweight extension to existing SCR architectures that conditions local visual features on their calibrated viewing rays and refines neighboring ray-conditioned features with a shallow local Transformer. We further formulate supervision directly in ray space and use sparse SfM points as metric anchors that link observations across views and camera models. Taken together, these components enable a single scene representation built from a panoramic mapping sequence to accurately localize query images captured with very different cameras, including pinhole, fisheye, and omnidirectional cameras.

We evaluate Ray-SCR across multiple SCR architectures and on datasets with pinhole, fisheye, and omnidirectional cameras. The proposed ray conditioning substantially improves generalization across cameras and leads to state-of-the-art localization performance in these challenging cross-camera settings.
The project page is available at \href{https://jaeminch0.github.io/ray-scr-generalizing-scene-coordinate-regression-across-camera-models/}{\textcolor{raspberry}{\textit{here}}}.

\end{abstract}

%% file: Section/1_Introduction.tex
\begin{figure}[t]
\centering
\includegraphics[width=\columnwidth]{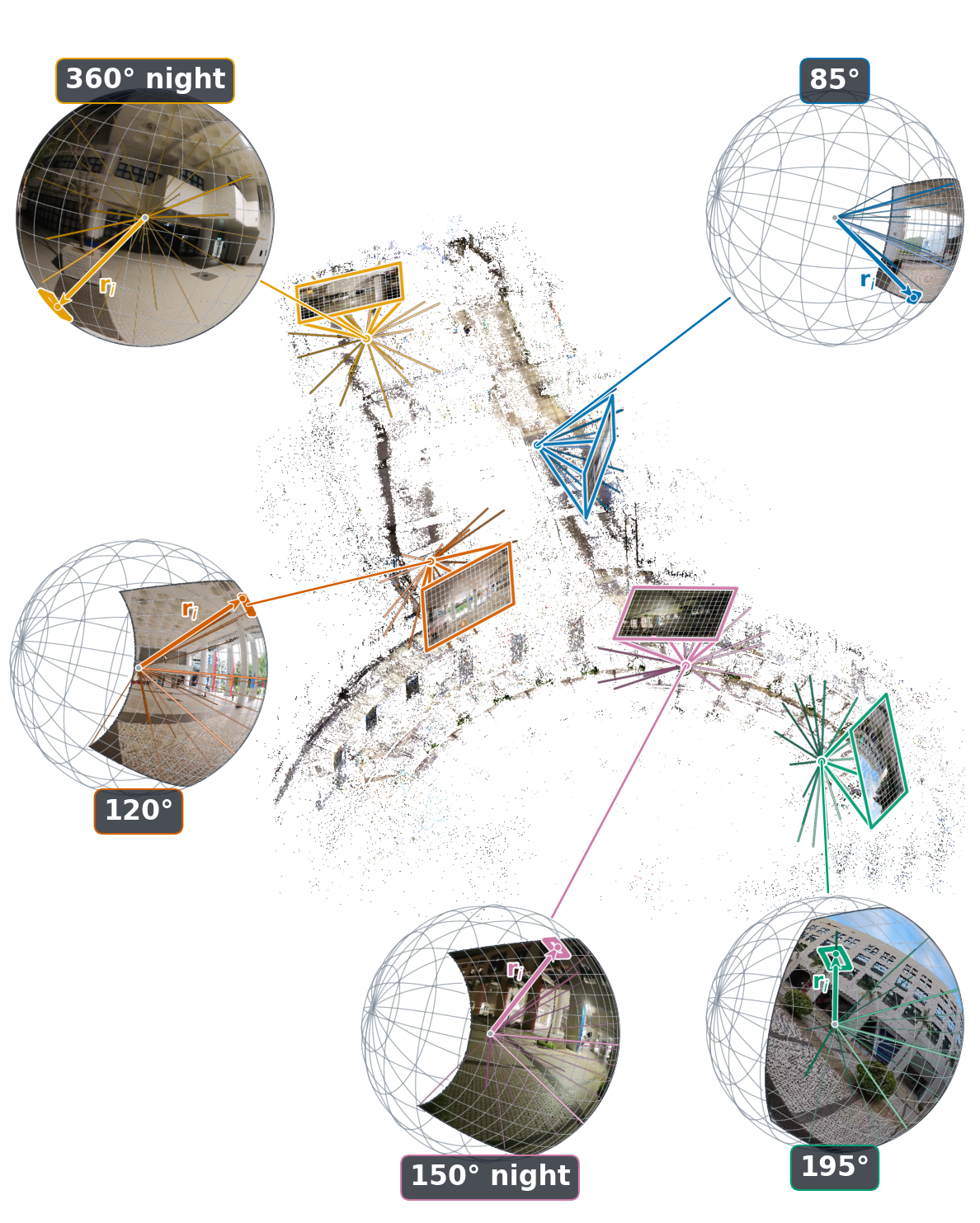}
\caption{\textbf{Cross-camera localization with Ray-SCR.} A scene mapped from panoramic images is queried using pinhole, fisheye, and $360^\circ$ cameras with substantially different fields of view and image distortions. Ray-SCR uses calibrated viewing rays associated with local image features to enable localization across these heterogeneous cameras.} 
\label{fig:ray_aceg_overview}
\vspace{-0.5cm}
\end{figure}

\section{Introduction}

Visual localization techniques allow to estimate the pose of a query image within a known scene. This process is central to many applications in robotics~\cite{rameau2022real}, mapping~\cite{schoenberger2016sfm}, and augmented reality~\cite{sarlin2022lamar}. Scene coordinate regression (SCR)~\cite{brachmann2017dsac, brachmann2023ace} approaches this task by learning, for each scene, a mapping from local image descriptors to 3D scene coordinates. Recently, SCR techniques have become particularly attractive for their fast training and inference, compact scene representation, and high localization accuracy.

Despite these advantages, existing SCR solutions have been designed primarily for perspective cameras, or rectified images, and do not explicitly account for the strong distortions and non-uniform image resolution induced by omnidirectional imaging systems. As a result, the same local scene structure can produce substantially different image features depending on the camera geometry, which limits the ability of SCR models to generalize across heterogeneous cameras.

To mitigate that limitation, one can train SCR with images rendered from different camera configurations, exposing the network to a wider range of distortions. However, SCR relies on local image features, while the distortion affecting each patch depends on its position in the camera field of view. Therefore, without access to the corresponding viewing direction, the network has no explicit information about where a local feature lies within the camera projection. As shown in \tableautorefname~\ref{tab:viewgen}, such augmentation significantly improves cross-camera localization, but remains insufficient on its own.

Building on this observation, we propose Ray-SCR, which augments each local visual feature with its corresponding calibrated camera ray, providing the network with the viewing direction associated with that observation. A shallow local transformer further refines neighboring features to capture projection-dependent local patterns. Beyond feature conditioning, we formulate the training objective directly in ray space, using the angular distance between predicted and observed viewing directions. Unlike image-plane reprojection, this measure is independent of the camera projection model and can therefore be applied consistently across pinhole, fisheye, and omnidirectional cameras. 
In addition, we exploit sparse SfM points as metric 3D anchors, linking observations across views and camera models to a common scene coordinate.

We validate Ray-SCR primarily on 360Loc~\cite{huang2024loc}, where scenes mapped from panoramic sequences are queried using five different camera configurations under both day and night conditions. Across several SCR backbones, Ray-SCR consistently improves cross-camera localization and outperforms existing cross-device methods, including CDG-Loc~\cite{wang2025cdgloc}. We further evaluate the approach on OmniScenes~\cite{kim2021piccolo}, where it surpasses scan-based localization methods~\cite{kim2024fully} while retaining a compact scene representation, and on Aachen~\cite{sattler2018benchmarking} to assess its behavior in a large-scale perspective setting with heterogeneous camera intrinsics.

\noindent
Our main contributions are:
\begin{itemize}
    \item We introduce Ray-SCR, which conditions local visual features on calibrated camera rays and jointly refines neighboring features to improve scene coordinate regression across heterogeneous cameras.
    \item  We formulate SCR supervision directly in ray space and complement it with sparse SfM anchors, providing metric 3D supervision while promoting consistent scene-coordinate predictions across views and camera models.
    \item We demonstrate the generality of Ray-SCR across several SCR architectures and three benchmarks, achieving state-of-the-art localization on 360Loc and OmniScenes, while substantially improving the corresponding SCR baseline on Aachen.
\end{itemize}

%% file: Section/2_Related_Work.tex
\section{Related Work}
In this section, we review scene coordinate regression and omnidirectional and cross-device localization.

\subsection{Scene Coordinate Regression}

Visual localization estimates the camera pose of a query image with respect to a pre-built map. A common approach is structure-based localization, which typically follows a hierarchical pipeline: it first retrieves database images depicting similar parts of the scene, then establishes local feature matches between the query and retrieved images to obtain 2D–3D correspondences with an SfM map, from which the camera pose is recovered geometrically~\cite{sattler2019understanding,sarlin2019coarse}.
Recently, FastForward~\cite{barroso2026fastforward} represents posed mapping images as a collection of features anchored in 3D space and localizes a query in a single feed-forward pass.

An alternative is to encode the scene implicitly in network weights, avoiding an explicit geometric map. For instance, absolute pose regression follows this paradigm by training a scene-specific network on images with known camera poses and directly regressing the pose of a new query from its learned representation~\cite{kendall2015posenet,shavit2021multi}. While such methods offer a compact and efficient localization pipeline, they lack the geometric grounding of correspondence-based estimation, leading to poor accuracy~\cite{sattler2019understanding}. 
Camera Ray Regression~\cite{zhang2025crr} instead predicts world-coordinate rays from image patches and recovers camera pose geometrically.

SCR bridges these two paradigms by encoding the scene in a network. Because this avoids storing an explicit scene representation at inference time, SCR has often been regarded as comparatively privacy-preserving, although recent work has challenged this assumption~\cite{nasypanyi2026seeingweights}.
Rather than predicting the camera pose directly, the network predicts, for each pixel, the corresponding 3D scene coordinate, thereby producing dense 2D–3D correspondences from which the camera pose is recovered with a robust PnP algorithm.
The seminal SCR approach of Shotton et al.\cite{shotton2013scene} learns to predict dense scene coordinates from RGB-D training data, providing 2D–3D correspondences that can be used for geometric pose estimation. To remove the reliance on ground-truth depth, DSAC*\cite{brachmann2022dsacstar} proposes to use a CNN to learn scene coordinates directly from posed RGB images. More recently, ACE~\cite{brachmann2023ace} then substantially accelerates this scene-specific training by combining a shared visual encoder with a lightweight coordinate regressor. Building on this pipeline, GLACE~\cite{wang2024glace} incorporates an image-level global encoding to improve scalability to larger scenes, while ACE-G~\cite{bruns2025aceg} introduces compact scene-specific map codes to reduce the mapping-to-query domain gap. FocusTune~\cite{nguyen2024focustune} further improves training by adapting the sampling of observations, while R-SCoRe~\cite{jiang2025rscore} targets robustness at larger scale and under challenging illumination. 
Despite these advances, existing SCR methods do not condition scene-coordinate prediction on the camera model. Camera geometry is only introduced afterward, through the reprojection objective and pose solver. Ray-SCR instead conditions the regressor directly on calibrated per-pixel viewing rays and uses the same ray representation for scoring and pose estimation, removing the assumption of a fixed perspective image plane.
Unlike predicting coordinate rays~\cite{zhang2025crr} and pose-dependent mapping rays~\cite{barroso2026fastforward}, Ray-SCR uses camera-coordinate rays to condition scene-coordinate prediction.
This makes a substantial difference in cross-camera settings, where the same scene may be observed through perspective, fisheye, or omnidirectional projections with very different pixel-to-ray mappings.

\subsection{Omnidirectional and Cross-Device Localization}

While hierarchical localization~\cite{sarlin2019coarse} is particularly effective on perspective imagery, its extension to cross-camera settings is not straightforward, as it requires retrieval strategies~\cite{shi2023panovpr} and keypoint descriptors~\cite{eder2020tangent,gava2023sphereglue} invariant to changes in projection model and field of view.

These limitations have led to omnidirectional localization methods that simplify the problem through stronger scene priors. PICCOLO and CPO localize panoramas directly in coloured point clouds through pose sampling and colour consistency~\cite{kim2021piccolo,kim2022cpo}, while LDL and FGPL further reduce the search space using dominant line directions and intersections~\cite{kim2023ldl,kim2024fully}. A complementary line of work replaces the 3D map with stronger indoor priors. For instance, LaLaLoc and LASER localize against floorplans~\cite{howard2022lalaloc++, min2022laser}, while SPVLoc uses a coarse semantic 3D model~\cite{gard2024spvloc}. While effective indoors, these approaches rely on strong assumptions about scene structure or map availability, limiting their applicability to more general environments. Moreover, they apply mostly to panoramic images and are often not suitable for fisheye or perspective cameras.

Cross-device localization, where the map is built with one device and queried with another, has so far been driven mainly by benchmarks. For instance, LaMAR~\cite{sarlin2022lamar} evaluates localization across phone and HoloLens~2 captures, while CroCoDL~\cite{blum2025crocodl} extends this setting to a legged robot. In both cases, however, the images remain perspective, and the domain gap mainly arises from changes in viewpoint, motion, and sensing platform. 360Loc~\cite{huang2024loc} instead explicitly studies changes in camera projection, mapping from $360^\circ$ images and querying with pinhole, fisheye, and $360^\circ$ cameras. Existing solutions typically handle such cross-device variations through input transformation or additional training data: 360Loc renders virtual perspective views to reuse conventional localization pipelines, SPR~\cite{zheng2025spr} learns scene-agnostic relative pose, and CDG-Loc~\cite{wang2025cdgloc} reconstructs scenes with 3DGS to synthesize additional training views. These approaches improve robustness to device changes, but do not provide a representation inherently shared across camera models. Ray-SCR instead uses viewing rays as a common geometric representation across projections.
Note that such invariance could also be obtained on the feature side, through spherical convolutions or distortion-aware descriptors. Doing so, however, means abandoning the strong pretrained encoders, for which no fisheye or 360 counterpart exists.

\begin{figure*}[t]
    \centering
    \includegraphics[width=\textwidth]{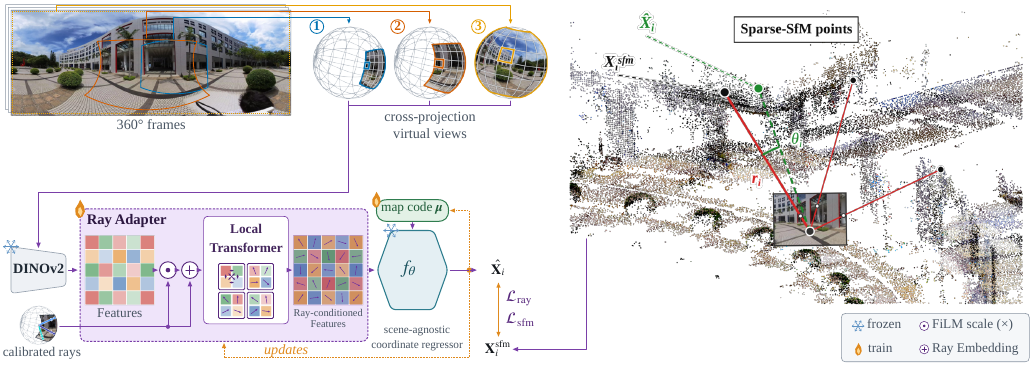}
    \caption{
    \textbf{Ray-SCR scene-training pipeline.}We first extract encoder features from the training images, optionally augmented with sampled virtual views. SfM points are projected into each view, and features whose angular distance $\theta_i$ from the corresponding viewing ray $\mathbf r_i$ is below a predefined threshold are retained in the training buffer. The Ray Adapter then injects camera-geometry information by encoding the associated viewing rays into the features and locally processing groups of neighboring features with a lightweight Transformer. The adapted features are subsequently passed through the standard SCR pipeline. Here, ACE-G is shown as an example, using a DINO encoder together with trainable scene-specific map codes.
    }
    \label{fig:pipeline}
    \vspace{-0.2cm}
\end{figure*}

%% file: Section/3_Methods_updated.tex
\section{Method}

\noindent\textbf{Background \& Rationale}
SCR encodes the geometry of a scene implicitly in the weights of a scene-specific network $f$, which maps an image $I$ to a field of 3D points. 
The resulting 2D-3D correspondences are passed to a robust PnP solver to recover the camera pose. 
Early SCR formulations trained $f$ entirely per scene~\cite{shotton2013scene,brachmann2022dsacstar}, while subsequent methods decompose it into a scene-agnostic feature extractor $f_{\mathrm{enc}}$ and a scene-specific prediction head $h$~\cite{brachmann2023ace,wang2024glace}. Given an image $I$, $f_{\mathrm{enc}}$ produces a set of patch embeddings
\begin{equation}
    \mathcal{F} = \{F_i \in \mathbb{R}^d\},
\end{equation}
where $F_i$ corresponds to the patch centered at pixel $\mathbf{u}_i$, and the head maps each embedding to a 3D point,
\begin{equation}
    \mathbf{y}_i = h(F_i; \mathbf{w}),
    \label{eq:scr_head}
\end{equation}
where $\mathbf{w}$ denotes the scene-specific parameters of $h$. More
recently, ACE-G~\cite{bruns2025aceg} pushes this decomposition one step
further and keeps $h$ fixed and shared across scenes, storing the scene in
a trainable map code $\boldsymbol{\mu}$ instead,
\begin{equation}
    \mathbf{y}_i = h(F_i, \boldsymbol{\mu}; \mathbf{w}).
    \label{eq:aceg_head}
\end{equation}
Ray-SCR acts before the prediction head, replacing $F_i$ by a ray-conditioned feature. It is therefore compatible with any of these backbones, regardless of how they store the scene.

While effective for perspective cameras, none of SCR formulations account for the viewing geometry,  treating local features identically regardless of their viewing direction. This becomes a problem for fisheye and panoramic images, where the appearance and sampling of a neighborhood depend strongly on its viewing direction. Note that undistortion, as shown in Table~\ref{tab:cubemap}, 
does not fully resolve this problem, as it introduces spatially varying resampling and stretching near the image boundaries.
Ray-SCR addresses this by associating each feature with its calibrated viewing ray (\ref{sec:problem_setting_ray_representation}). The ray modulates the feature, and a lightweight local transformer then adapts each neighborhood into a representation shared across perspective, fisheye, and panoramic cameras (Sec.~\ref{sec:ray_conditioned_scr}). We further use sparse SfM points, already available from scene reconstruction, as direct 3D anchors (Sec.~\ref{sec:ray_space_scene_mapping}). Since a point is observed from several viewpoints and camera models, these anchors push different observations of the same structure toward a consistent 3D coordinate.

\subsection{Problem Setting and Ray Representation}
\label{sec:problem_setting_ray_representation}

We consider a scene $\mathcal{M}$ represented by a posed image sequence,
$
\mathcal{M}={(I_j,\mathbf{T}^{j}_{w\leftarrow c})}{j=1}^{N},
$
where $\mathbf{T}^{j}_{w\leftarrow c}$ denotes the camera-to-world pose associated with image $I_j$. Given a query image $I_q$, potentially captured using a different camera model, our goal is to estimate its pose in the scene coordinate defined by $\mathcal{M}$. For each pixel location $\mathbf{u}_i=(u_i,v_i)^\top$, the corresponding camera calibration $\kappa$ defines a unit viewing ray
\begin{equation}
\mathbf{r}_i = \rho_{\kappa}(\mathbf{u}_i) \in \mathbb{S}^{2},
\label{eq:calibrated_camera_ray}
\end{equation}
where $\rho_{\kappa}$ denotes the inverse projection model of camera $\kappa$. Since the ray field depends only on the camera calibration and the feature-grid resolution, it is computed once for each camera configuration. Each feature $F_i$ is then directly associated with the ray $\mathbf{r}_i$ passing through the center of its corresponding image patch.

\subsection{Training Buffer Construction}
\label{sec:training_buffer}

Following ACE\cite{brachmann2023ace}, we precompute encoder features in a training buffer, but sample 8×8 spatial tiles rather than individual features so that neighboring ray-conditioned features can be jointly processed by the lightweight Transformer described in Sec.~\ref{sec:ray_conditioned_scr}.
Tiles are shuffled across the buffer before training, retaining the randomized training strategy of ACE. Since each optimization batch contains several thousand features, preserving groups of only $64$ neighboring features has little effect on the overall sampling diversity.

Additionally, we optionally sample virtual perspective and fisheye views from the $360^\circ$ images to populate the training buffer. For simplicity, we use the known camera calibrations provided by the dataset and randomly sample camera orientations around the original camera center. We then generate each virtual view by reprojecting the panoramic image according to the sampled orientation and the corresponding camera model, following the procedure described in~\cite{huang2024loc}. 

Finally, we associate sparse SfM observations with the feature grid. For each mapping image, we retain the top $N_{\mathrm{sfm}}$ most reliable SfM observations, ranked by increasing reprojection error, and assign each observation to the closest feature according to angular distance.
An association is accepted only when their angular distance is below $\tau_{\mathrm{sfm}}=2^\circ$; otherwise the feature is marked as having no SfM supervision. As a result, only a sparse subset of feature locations, approximately $5\%$ in our setup, receives a direct 3D target. 
Yet this sparse supervision provides geometric links between observations across different views and camera models.

\subsection{Ray-Conditioned Feature Adaptation}
\label{sec:ray_conditioned_scr}

Given a feature $F_i$ sampled from the training buffer and its corresponding calibrated unit ray $\mathbf{r}_i$, we encode the viewing direction using a Fourier feature mapping $\gamma(\cdot)$,
\begin{equation}
\mathbf{q}_i = \gamma(\mathbf{r}_i) \in \mathbb{R}^{64},
\label{eq:ray_fourier_encoding}
\end{equation}
where $\mathbf{q}_i$ denotes the resulting ray embedding.
This ray embedding is injected into the visual feature through two complementary branches. 
First, a lightweight MLP
$g_{\mathrm{film}}:\mathbb{R}^{64}\rightarrow\mathbb{R}^{2d}$
predicts feature-wise FiLM~\cite{perez2018film} scale and shift parameters,
\begin{equation}
[\boldsymbol{\alpha}_i,\boldsymbol{\beta}_i]
= g_{\mathrm{film}}(\mathbf{q}_i).
\end{equation}
In parallel, a learned projection
$g_{\mathrm{ray}}:\mathbb{R}^{64}\rightarrow\mathbb{R}^{d}$
maps the ray embedding directly into the visual feature space,
\begin{equation}
\mathbf{e}^{\mathrm{ray}}_i
= g_{\mathrm{ray}}(\mathbf{q}_i).
\end{equation}
The first branch modulates the visual feature, while the second provides an additive ray-dependent embedding. Together, they yield
\begin{equation}
F_i^{\mathrm{ray}}
=
F_i \odot (1+\boldsymbol{\alpha}_i)
+\boldsymbol{\beta}_i
+\mathbf{e}^{\mathrm{ray}}_i .
\label{eq:ray_feature_conditioning}
\end{equation}

\begin{figure}
    \centering
    \includegraphics[width=1\linewidth]{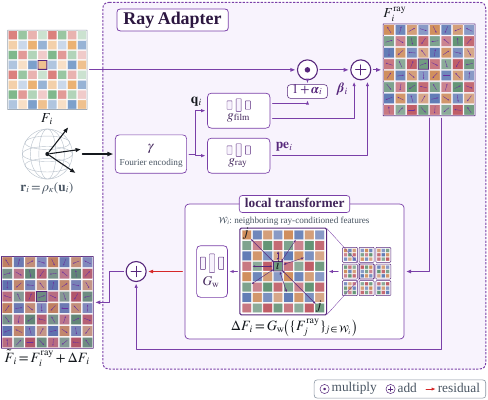}
    \caption{\textbf{Ray Adapter Architecture.}  Feature tiles sampled from the training buffer are paired viewing rays. Each ray is encoded and used to condition the visual feature through FiLM modulation and an additive ray embedding. The resulting features are then jointly refined within each local tile using a local Transformer before being passed to the SCR prediction head.} 
    \label{fig:adapter}
    \vspace{-0.2cm}
\end{figure}

Applying this conditioning independently to all $64$ features within a sampled $8\times8$ neighborhood yields a ray-conditioned feature tile. We then process this tile with a shallow local transformer, allowing neighboring features to interact and thereby account for projection-dependent variations in local feature patterns. The transformer operates only within each tile, keeping the additional computation lightweight and preserving compatibility with existing SCR prediction heads(Fig.~\ref{fig:adapter}). At inference, the feature map is processed in non-overlapping $8\times8$ tiles with a stride of $8$.

\subsection{Training Objective}
\label{sec:ray_space_scene_mapping}

Ray-SCR is trained using two complementary losses. Given a predicted scene coordinate $\widehat{\mathbf{X}}i$ and the known mapping pose $\mathbf{T}_{w\leftarrow c}=(\mathbf{R}_{w\leftarrow c},\mathbf{t}_w)$, where $\mathbf{R}_{w\leftarrow c}$ denotes the camera-to-world rotation and $\mathbf{t}_w$ the camera center expressed in world coordinates, the viewing direction induced by the prediction is

\begin{equation}
\widehat{\mathbf r}_i =
\frac{
\mathbf{R}_{w\leftarrow c}^{\top}
(\widehat{\mathbf X}_i-\mathbf t_w)
}{
\left|
\mathbf{R}_{w\leftarrow c}^{\top}
(\widehat{\mathbf X}_i-\mathbf t_w)
\right|_2
}.
\end{equation}
We supervise its alignment with the calibrated observation ray using the angular residual
\begin{equation}
\theta_i =
\operatorname{atan2}\left(
||\mathbf r_i\times\widehat{\mathbf r}_i||_2,
\mathbf r_i^\top\widehat{\mathbf r}_i
\right),
\end{equation}
yielding the loss
\begin{equation}
\mathcal{L}_{\mathrm{ray}}
=
\frac{1}{|\mathcal V|}
\sum_{i\in\mathcal V}\theta_i .
\end{equation}

We complement the dense ray-space supervision with sparse SfM anchors, which provide direct metric targets and link observations of the same scene structure across different views and camera models.
At locations with an associated SfM point $\mathbf X_i^{\mathrm{sfm}}$, we therefore apply
\begin{equation}
\mathcal{L}_{\mathrm{sfm}}
=
\frac{1}{|\mathcal S|}
\sum_{i\in\mathcal S}
w_i,
\operatorname{SmoothL1}
\left(
\widehat{\mathbf X}_i-\mathbf X_i^{\mathrm{sfm}}
\right),
\end{equation}
where $w_i$ denotes the SfM confidence. The final mapping objective is
\begin{equation}
\mathcal{L}_{\mathrm{map}}
=
\mathcal{L}_{\mathrm{ray}}
+
\lambda_{\mathrm{sfm}}\mathcal{L}_{\mathrm{sfm}}.
\end{equation}

\subsection{Pose Estimation} 
\label{sec:pose_estimation_calibrated_rays}
Each image location is first mapped through the given camera model to a unit bearing vector on the viewing sphere. This bearing is paired with the 3D coordinate predicted by the SCR model, forming a set of bearing–3D correspondences. We estimate the camera pose using a bearing-vector P3P solver within RANSAC, as implemented in OpenGV~\cite{huang2025opengv}, and then refine the resulting pose with robust nonlinear least squares by minimizing the angular error between the observed bearing vectors and the directions of the predicted 3D points under the estimated pose.

%% file: Section/4_Experiments.tex
\section{Experiments}
\input{Table/tab1_day}
\input{Table/tab2_night}
\input{Table/tab3_scr}

\subsection{Experimental Setup}
\label{sec:exp_setup}

\noindent\textbf{Datasets.}
We evaluate Ray-SCR on three complementary benchmarks covering different scales, camera configurations, and localization settings. 360Loc~\cite{huang2024loc}, our primary benchmark, covers campus-scale indoor and outdoor environments with symmetric structures, repetitive appearance, and dynamic pedestrians. Each scene is mapped from a posed $360^\circ$ sequence and queried using pinhole, fisheye, and $360^\circ$ cameras under day and night conditions. We further evaluate on OmniScenes~\cite{kim2021piccolo}, a room-scale indoor benchmark with panoramas captured from handheld and mobile robot-mounted cameras before and after scene rearrangement; we use its joint-space setting to evaluate localization between handheld and robot-mounted captures under scene rearrangement. Aachen~\cite{sattler2018benchmarking} complements these experiments with a city-scale outdoor setting, where mapping and query images exhibit varied camera intrinsics and fields of view.

\noindent\textbf{Implementation Details.}
We implement Ray-SCR in PyTorch and optimize its trainable components with mixed-precision AdamW. On 360Loc, mapping images are resized to a maximum long edge of 1024 pixels, and each scene is trained for 70{,}000 iterations on NVIDIA RTX A5000 GPUs. For Ray-ACEG, we use a training buffer size of 9M and learning rates of $1\mathrm{e}{-4}$ for the Ray Adapter and $3\mathrm{e}{-4}$ for the scene-specific map code. The ray reprojection and SfM anchor losses use equal unit weights; the latter uses Smooth-$L_1$ with $\beta=0.25$ and at most 4096 SfM observations per mapping image.
The Ray adapter uses 10 Fourier frequencies and a low-rank width of 64, followed by a single-layer, 4-head local Transformer with 256-dimensional tokens operating on $8{\times}8$ windows. The final projections of the FiLM, additive ray embedding, and Transformer residual branches are zero-initialized, so that the adapter initially preserves the pretrained features.
Settings for the other SCR backbones, OmniScenes, and Aachen are provided in the supplementary material.

\subsection{Cross-Camera Localization}
On 360Loc, we compare Ray-ACEG with the absolute pose regression baselines PoseNet (PN)~\cite{kendall2015posenet} and MS-Transformer (MS-T)~\cite{shavit2021multi}, their virtual-camera variants PN-VC2 and MS-T-VC2~\cite{huang2024loc}, and the cross-device localization method CDG-Loc~\cite{wang2025cdgloc}. We use the baseline results reported by CDG-Loc under the same evaluation protocol. For each scene and query camera, we report the median translation and rotation errors.

\noindent\textbf{Day and night localization.}
For the main comparison, we instantiate Ray-SCR with ACE-G, using a single scene model to localize all five query cameras. Following CDG-Loc\cite{wang2025cdgloc}, each scene map additionally includes 10--20 nighttime frames. Tables~\ref{tab:360loc_day} and~\ref{tab:360loc_night} show that Ray-ACEG achieves the lowest rotation error for every camera type and scene under both illumination conditions, as well as the lowest scene-average translation error in six of eight settings. Compared with the previous best method, CDG-Loc, Ray-ACEG reduces the average rotation error by more than $2\times$ across the eight scene--illumination settings. The largest gains are observed for narrow-FoV pinhole queries, with the advantage retained at night.

\begin{figure}[t]
\centering
\includegraphics[width=\columnwidth]{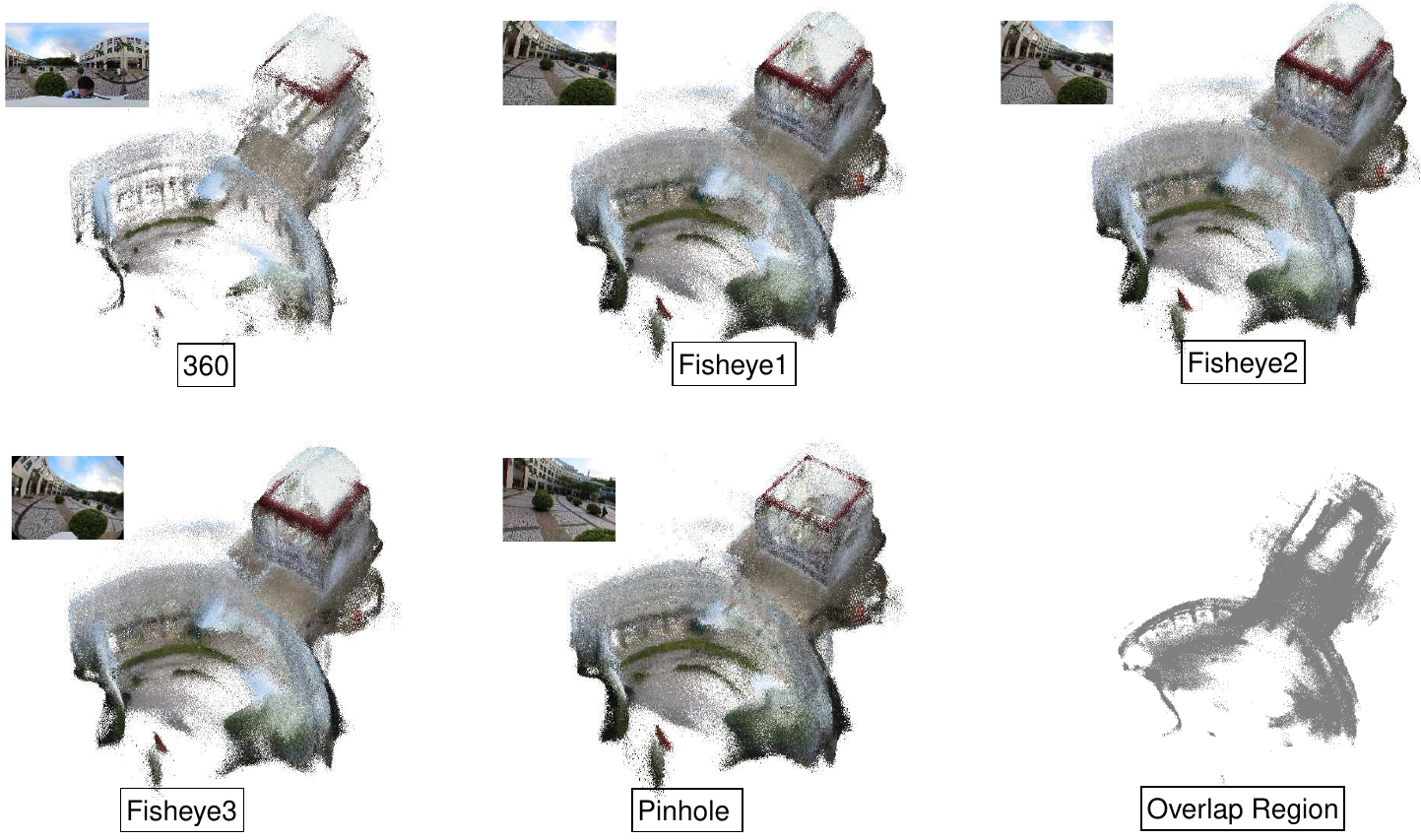}
\caption{\textbf{Cross-camera consistency of predicted scene coordinates on Piatrium-Day.} Ray-SCR produces substantially overlapping 3D scene points from all five camera types.}
\label{fig:qual_cross_camera_scene_coords}
\vspace{-0.2cm}
\end{figure}

\noindent\textbf{Compatibility across SCR architectures.}
We generalize Ray-SCR across SCR backbones by applying the same ray conditioning, training objective, and pose estimation to DSAC*~\cite{brachmann2022dsacstar}, ACE~\cite{brachmann2023ace}, and GLACE~\cite{wang2024glace} (Table~\ref{tab:360loc_scr_day}). Unlike ACE-G, these backbones encode each scene in a scene-specific prediction head, with DSAC* additionally learning its encoder per scene. Despite this architectural difference, all three Ray-SCR variants localize the five camera types and outperform the original baselines in most 360Loc Day settings. Among them, Ray-GLACE provides the strongest results, indicating that the benefits of Ray-SCR are not tied to the ACE-G architecture.

\subsection{Omnidirectional and Perspective Localization}

\noindent\textbf{OmniScenes.}
We compare against LDL~\cite{kim2023ldl} and FGPL~\cite{kim2024fully}, and retrieval-and-matching pipelines combining CosPlace (CP)~\cite{berton2022rethinking} or SFRS~\cite{ge2020self} with LightGlue (LG)~\cite{lindenberger2023lightglue}, GlueStick (GS)~\cite{pautrat2023gluestick}, or GoMatch (GM)~\cite{zhou2022geometry}; all use explicit 3D maps. Since OmniScenes provides no mapping split, we instead learn an implicit neural map for each space from posed ``unchanged'' panoramas and localize ``changed'' queries through predicted 2D--3D correspondences on the joint seven-space Extreme split, where all seven spaces form a single localization problem.

\begin{figure}[t]
\centering
\includegraphics[width=\columnwidth]{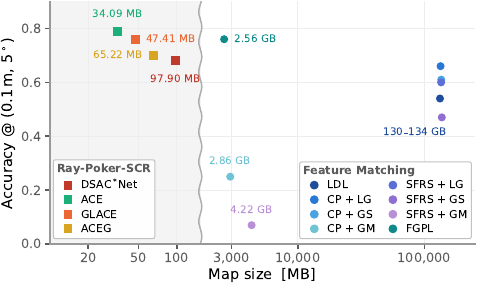}
\caption{\textbf{Map size and localization accuracy on OmniScenes.} Explicit-map methods (circles) use a pre-captured 3D point-cloud map at query time, whereas Ray-Poker-SCR (squares) represents each scene implicitly in network weights. Ray-Poker-ACE achieves the highest accuracy with $77\times$ less storage than
FGPL~\cite{kim2024fully}.}
\label{fig:mapsize_acc}
\vspace{-0.2cm}
\end{figure}

To deal with this multi-space setting, we follow ACE's Poker strategy~\cite{brachmann2023ace}: Ray-Poker-ACE estimates poses with all seven per-space regressors, selects the hypothesis with the strongest geometric support, and refines it. The same procedure is used for GLACE, DSAC*, and ACE-G.
Unlike LDL and FGPL, which process perspective crops rendered from the panorama, all four Ray-Poker-SCR variants directly consume each $360^\circ$ query without cubemap conversion. Across the seven spaces, the four Ray-Poker-SCR variants achieve competitive recall accuracy. Among them, Ray-Poker-ACE improves recall by $3.9\%$ over FGPL while using $77\times$ less scene-specific storage (Fig.~\ref{fig:mapsize_acc}).

\noindent\textbf{Aachen.}
We train Ray-ACE and Ray-GLACE directly on calibrated perspective images, following the ACE multi-map setup~\cite{brachmann2023ace} and the GLACE single-map setup~\cite{wang2024glace}, respectively. For Ray-ACE, we share a single Ray Adapter across all heads and select the pose with the greatest inlier support at inference. Table~\ref{tab:aachen} shows that both Ray-SCR variants outperform their respective baselines at all three pose thresholds. Ray-SCR maintains its advantage, with Ray-ACE ($N=50$) achieving the highest localization accuracy among the evaluated configurations. These results support conditioning local visual features on calibrated viewing rays for city-scale localization across varying camera intrinsics and fields of view.

\input{Table/tab7_aachen}

\subsection{Ablation Experiments}

\input{Table/tab6_viewgen}
\noindent\textbf{Virtual-view augmentation.}
Ray-SCR augments the posed $360^\circ$ mapping sequence with calibrated pinhole, fisheye, and $360^\circ$ views (Fig.~\ref{fig:pipeline}). 
As shown in Table~\ref{tab:viewgen}, expanding Ray-ACEG training from only $360^\circ$ views to all three camera families reduces both errors.
Standard ACE-G also benefits substantially from camera-diverse augmentation, but Ray-ACEG remains more accurate under every matched configuration, showing further gains from explicit ray conditioning.

\input{Table/tab8_cubemap}
\vspace{0.2cm}
\noindent\textbf{Cubemap-based SCR.}
A practical alternative for heterogeneous cameras is to resample calibrated images into perspective cubemap faces and apply standard SCR on these pinhole views.
We train Cubemap-ACE and Cubemap-ACEG on six $512{\times}512$ perspective faces per mapping panorama, each with a $90^\circ$ field of view. Queries use the same conversion, discarding faces below $35\%$ valid coverage, whereas Ray-SCR directly processes native queries (Table~\ref{tab:cubemap}).
Cubemap-ACE outperforms Ray-ACE on $360^\circ$ queries but has larger translation errors on pinhole and fisheye queries. In contrast, Ray-ACEG outperforms Cubemap-ACEG in both translation and rotation for every query camera, highlighting the limits of perspective resampling alone.

\vspace{0.2cm}
\noindent\textbf{Component analysis.}
\input{Table/tab5_ray_ladder}
Table~\ref{tab:ray_ladder} evaluates the local Transformer and the two training losses on Concourse under the same training data and budget. 
Removing $\mathcal{L}_{\mathrm{ray}}$ leads to the largest increases in translation and rotation errors under both day and night conditions, highlighting the importance of dense angular supervision.
Removing $\mathcal{L}_{\mathrm{sfm}}$ also degrades performance, showing the benefit of metric anchors that link observations across views to consistent scene coordinates. The full model performs best under both day and night conditions.
Additional experiments investigating the impact of the number of input tokens in the local Transformer are provided in the supplementary material.

%% file: Table/tab1_day.tex
% Baselines transcribed verbatim from CDG-Loc (SOTA) Tabs. 2-3; ours rounded to one decimal.
\begin{table*}[t]
\centering
\caption{\textbf{Cross-camera localization on 360Loc Day.} Median translation/rotation errors (m/$^\circ$). Best results are shown in \textbf{bold} and second-best results are \underline{underlined}.}
\label{tab:360loc_day}
\vspace{1mm}
\resizebox{\textwidth}{!}{
\begin{tabular}{c|l|cccccc|cccccc}
\hline
 & Scene & \multicolumn{6}{c|}{Atrium Day} & \multicolumn{6}{c}{Concourse Day} \\
\cline{2-14}
 & Camera & Pinhole & Fisheye1 & Fisheye2 & Fisheye3 & 360 & Average & Pinhole & Fisheye1 & Fisheye2 & Fisheye3 & 360 & Average \\
\hline
\multirow{5}{*}{\rotatebox[origin=c]{90}{APR}} & PN~\cite{kendall2015posenet} & 15.5/85.5 & 11.8/79.9 & 11.3/80.4 & 10.1/79.8 & 4.5/82.6 & 10.6/81.6 & 18.6/95.0 & 10.7/88.7 & 10.1/87.7 & 7.7/85.9 & 2.6/39.0 & 9.9/79.3 \\
 & MS-T~\cite{shavit2021multi} & 20.1/77.7 & 14.5/73.7 & 14.0/71.7 & 12.5/65.6 & 13.2/54.7 & 14.9/68.7 & 17.0/69.1 & 11.0/60.8 & 10.5/59.6 & 9.9/63.4 & 4.9/34.6 & 10.7/57.5 \\
 & PN-VC2~\cite{huang2024loc} & 11.5/31.2 & 9.1/19.5 & 8.9/19.3 & 8.4/18.1 & 7.7/17.1 & 9.1/21.0 & 7.0/18.9 & 4.7/11.5 & 4.4/11.2 & 3.9/9.9 & 3.2/6.9 & 4.6/11.7 \\
 & MS-T-VC2~\cite{huang2024loc} & 10.0/46.7 & 6.3/36.3 & 6.2/36.5 & 4.9/28.1 & 5.2/47.6 & 6.5/39.0 & 3.5/19.6 & 1.9/12.7 & 1.8/12.8 & 1.7/12.7 & 1.6/13.5 & 2.1/14.3 \\
 & CDG-Loc~\cite{wang2025cdgloc} & \underline{5.2}/\underline{21.8} & \underline{2.2}/\underline{12.7} & \underline{2.1}/\underline{12.3} & \underline{1.8}/\underline{11.5} & \underline{1.6}/\underline{14.1} & \underline{2.6}/\underline{14.5} & \underline{2.7}/\underline{9.7} & \underline{1.5}/\underline{5.7} & \underline{1.3}/\underline{5.5} & \underline{1.1}/\underline{5.4} & \underline{1.0}/\underline{5.7} & \underline{1.5}/\underline{6.4} \\
\cline{1-14}
\multirow{1}{*}{SCR} & Ray-ACEG (Ours) & \textbf{1.9}/\textbf{4.3} & \textbf{1.2}/\textbf{2.9} & \textbf{1.2}/\textbf{3.0} & \textbf{1.1}/\textbf{2.7} & \textbf{0.9}/\textbf{2.3} & \textbf{1.3}/\textbf{3.0} & \textbf{1.4}/\textbf{6.0} & \textbf{0.9}/\textbf{4.4} & \textbf{1.0}/\textbf{4.5} & \textbf{0.8}/\textbf{4.1} & \textbf{0.7}/\textbf{3.5} & \textbf{1.0}/\textbf{4.5} \\
\hline
 & Scene & \multicolumn{6}{c|}{Hall Day} & \multicolumn{6}{c}{Piatrium Day} \\
\cline{2-14}
 & Camera & Pinhole & Fisheye1 & Fisheye2 & Fisheye3 & 360 & Average & Pinhole & Fisheye1 & Fisheye2 & Fisheye3 & 360 & Average \\
\hline
\multirow{5}{*}{\rotatebox[origin=c]{90}{APR}} & PN~\cite{kendall2015posenet} & 19.8/90.8 & 11.1/95.7 & 10.4/95.9 & 7.5/92.3 & 2.7/97.3 & 10.1/96.1 & 21.3/88.0 & 13.8/86.5 & 13.0/87.1 & 10.9/84.4 & 5.7/69.8 & 12.9/83.2 \\
 & MS-T~\cite{shavit2021multi} & 21.8/96.0 & 14.1/94.0 & 13.5/92.7 & 11.2/91.7 & 11.4/88.5 & 14.4/92.6 & 30.0/78.5 & 23.4/71.3 & 22.7/72.7 & 19.3/72.6 & 11.3/52.0 & 21.3/69.4 \\
 & PN-VC2~\cite{huang2024loc} & 7.0/26.4 & 4.4/15.5 & 4.3/14.7 & 3.9/14.2 & 3.2/12.2 & 4.6/16.6 & 10.4/24.5 & 6.9/13.3 & 6.8/12.8 & 6.8/11.5 & 5.0/\underline{9.8} & 7.2/14.4 \\
 & MS-T-VC2~\cite{huang2024loc} & 4.0/36.9 & 2.2/25.8 & 2.1/25.2 & 1.8/26.2 & 1.7/26.9 & 2.4/28.2 & 8.1/32.7 & 4.0/24.5 & 3.9/23.6 & 3.8/24.1 & 3.3/22.1 & 4.6/25.4 \\
 & CDG-Loc~\cite{wang2025cdgloc} & \underline{2.3}/\underline{17.4} & \textbf{1.0}/\underline{8.9} & \textbf{1.0}/\underline{8.6} & \textbf{0.9}/\underline{8.4} & \textbf{0.7}/\underline{7.5} & \underline{1.2}/\underline{10.2} & \underline{4.4}/\underline{18.0} & \textbf{2.5}/\underline{11.3} & \textbf{2.4}/\underline{10.5} & \textbf{2.4}/\underline{10.1} & \textbf{2.3}/10.1 & \textbf{2.8}/\underline{12.0} \\
\cline{1-14}
\multirow{1}{*}{SCR} & Ray-ACEG (Ours) & \textbf{1.5}/\textbf{6.8} & \underline{1.1}/\textbf{4.4} & \underline{1.0}/\textbf{4.2} & \underline{0.9}/\textbf{4.4} & \underline{0.9}/\textbf{4.1} & \textbf{1.1}/\textbf{4.8} & \textbf{4.2}/\textbf{8.5} & \underline{2.8}/\textbf{5.1} & \underline{2.6}/\textbf{4.9} & \underline{2.7}/\textbf{5.2} & \underline{2.4}/\textbf{5.7} & \underline{2.9}/\textbf{5.9} \\
\hline
\end{tabular}
}
\end{table*}

%% file: Table/tab2_night.tex
% Baselines transcribed verbatim from CDG-Loc (SOTA) Tabs. 2-3; ours rounded to one decimal.
\begin{table*}[t]
\centering
\caption{\textbf{Cross-camera localization on 360Loc Night.} Median translation/rotation errors (m/$^\circ$). Best results are shown in \textbf{bold} and second-best results are \underline{underlined}.}
\label{tab:360loc_night}
\vspace{1mm}
\resizebox{\textwidth}{!}{
\begin{tabular}{c|l|cccccc|cccccc}
\hline
 & Scene & \multicolumn{6}{c|}{Atrium Night} & \multicolumn{6}{c}{Concourse Night} \\
\cline{2-14}
 & Camera & Pinhole & Fisheye1 & Fisheye2 & Fisheye3 & 360 & Average & Pinhole & Fisheye1 & Fisheye2 & Fisheye3 & 360 & Average \\
\hline
\multirow{5}{*}{\rotatebox[origin=c]{90}{APR}} & PN~\cite{kendall2015posenet} & 19.8/90.8 & 16.4/87.3 & 16.1/86.7 & 15.7/83.5 & 8.7/22.8 & 15.3/74.2 & 19.8/96.2 & 13.3/85.5 & 13.0/88.4 & 11.1/87.5 & 5.5/45.6 & 12.5/80.6 \\
 & MS-T~\cite{shavit2021multi} & 24.8/84.7 & 17.8/80.5 & 17.0/80.1 & 15.1/80.0 & 5.0/46.3 & 15.9/74.3 & 22.1/74.7 & 17.7/69.5 & 17.0/68.3 & 17.2/72.7 & 8.4/57.3 & 16.5/68.5 \\
 & PN-VC2~\cite{huang2024loc} & 13.6/50.2 & 11.5/38.1 & 11.6/38.3 & 11.4/37.8 & 9.7/45.7 & 11.6/42.0 & 10.7/25.4 & 8.3/16.4 & 8.4/15.7 & 8.1/14.7 & 5.5/13.5 & 8.2/17.1 \\
 & MS-T-VC2~\cite{huang2024loc} & 13.5/70.6 & 9.1/64.1 & 9.0/63.8 & 8.6/64.4 & 7.3/61.5 & 9.5/64.9 & 5.5/28.7 & 4.0/20.1 & 3.9/20.3 & 3.5/20.9 & 3.4/23.7 & 4.1/22.7 \\
 & CDG-Loc~\cite{wang2025cdgloc} & \underline{4.6}/\underline{25.3} & \textbf{1.8}/\underline{11.0} & \textbf{1.7}/\underline{10.1} & \textbf{1.6}/\underline{10.1} & \textbf{1.4}/\underline{7.3} & \textbf{2.2}/\underline{12.8} & \underline{3.3}/\underline{11.4} & \underline{1.7}/\underline{6.6} & \underline{1.6}/\underline{6.3} & \underline{1.4}/\underline{6.9} & \textbf{1.1}/\underline{6.1} & \underline{1.8}/\underline{7.5} \\
\cline{1-14}
\multirow{1}{*}{SCR} & Ray-ACEG (Ours) & \textbf{4.2}/\textbf{10.9} & \underline{2.4}/\textbf{6.5} & \underline{2.3}/\textbf{6.2} & \underline{2.2}/\textbf{6.5} & \underline{2.1}/\textbf{6.0} & \underline{2.7}/\textbf{7.2} & \textbf{1.7}/\textbf{8.0} & \textbf{1.2}/\textbf{5.8} & \textbf{1.2}/\textbf{5.9} & \textbf{1.2}/\textbf{5.5} & \underline{1.2}/\textbf{5.0} & \textbf{1.3}/\textbf{6.0} \\
\hline
 & Scene & \multicolumn{6}{c|}{Hall Night} & \multicolumn{6}{c}{Piatrium Night} \\
\cline{2-14}
 & Camera & Pinhole & Fisheye1 & Fisheye2 & Fisheye3 & 360 & Average & Pinhole & Fisheye1 & Fisheye2 & Fisheye3 & 360 & Average \\
\hline
\multirow{5}{*}{\rotatebox[origin=c]{90}{APR}} & PN~\cite{kendall2015posenet} & 22.8/95.6 & 18.3/93.8 & 17.9/93.6 & 16.8/91.3 & 10.9/100.1 & 17.3/94.9 & 33.2/89.7 & 29.3/89.6 & 28.8/90.4 & 29.2/89.8 & 30.6/98.5 & 30.2/91.6 \\
 & MS-T~\cite{shavit2021multi} & 32.1/96.1 & 29.9/94.2 & 29.7/93.6 & 31.8/93.1 & 22.3/94.4 & 29.2/94.3 & 37.7/91.3 & 37.3/89.1 & 36.2/89.0 & 35.7/85.6 & 37.2/83.0 & 36.8/87.6 \\
 & PN-VC2~\cite{huang2024loc} & 13.1/56.1 & 10.5/43.7 & 10.1/43.0 & 9.7/40.5 & 7.5/34.7 & 10.2/43.6 & 15.8/51.9 & 13.2/34.9 & 12.8/34.8 & 12.5/32.6 & 10.6/30.9 & 13.0/37.0 \\
 & MS-T-VC2~\cite{huang2024loc} & 8.9/73.1 & 5.6/67.6 & 5.3/66.3 & 4.8/68.9 & 3.8/67.2 & 5.7/68.6 & 16.0/63.9 & 13.0/53.8 & 12.4/53.7 & 11.9/49.9 & 11.9/49.9 & 13.0/54.2 \\
 & CDG-Loc~\cite{wang2025cdgloc} & \underline{4.2}/\underline{32.5} & \underline{2.2}/\underline{18.5} & \underline{2.1}/\underline{17.5} & \underline{1.9}/\underline{15.7} & \underline{1.5}/\underline{18.0} & \underline{2.4}/\underline{20.4} & \underline{7.2}/\underline{27.9} & \underline{4.6}/\underline{16.1} & \underline{4.3}/\underline{16.1} & \underline{3.7}/\underline{14.1} & \textbf{2.6}/\underline{9.6} & \underline{4.5}/\underline{16.8} \\
\cline{1-14}
\multirow{1}{*}{SCR} & Ray-ACEG (Ours) & \textbf{2.6}/\textbf{12.1} & \textbf{1.6}/\textbf{6.8} & \textbf{1.6}/\textbf{7.3} & \textbf{1.6}/\textbf{7.0} & \textbf{1.1}/\textbf{5.2} & \textbf{1.7}/\textbf{7.7} & \textbf{6.2}/\textbf{13.2} & \textbf{3.3}/\textbf{7.1} & \textbf{3.2}/\textbf{7.1} & \textbf{3.1}/\textbf{7.2} & \underline{2.7}/\textbf{8.5} & \textbf{3.7}/\textbf{8.6} \\
\hline
\end{tabular}
}
\end{table*}

%% file: Table/tab3_scr.tex
\begin{table}[t]
\centering
\caption{\textbf{Compatibility across SCR models on 360Loc Day.} In addition to the Ray-ACEG results, we instantiate Ray-SCR with DSAC*, ACE, and GLACE. All four models are evaluated on the five query camera types. Values are median translation/rotation errors (m/$^\circ$), with the best results shown in \textbf{bold}.}
\label{tab:360loc_scr_day}
\vspace{1mm}
\resizebox{\columnwidth}{!}{%
\begin{tabular}{l|cccccc}
\hline
\multicolumn{7}{c}{Atrium Day} \\
\hline
Camera & Pinhole & Fisheye1 & Fisheye2 & Fisheye3 & 360 & Avg \\
\hline
\hline
Ray-DSAC$^*$Net & 14.2/28.1 & 5.0/10.8 & 4.5/9.8 & 5.8/13.1 & 7.2/12.1 & 7.3/14.8 \\
Ray-ACE & 13.5/23.3 & 5.2/11.5 & 4.9/11.8 & 6.0/13.5 & 7.9/8.1 & 7.5/13.6 \\
Ray-GLACE & 7.5/18.1 & 2.8/6.8 & 2.7/7.0 & 3.7/11.6 & 4.3/8.0 & 4.2/10.3 \\
Ray-ACEG  & \textbf{1.9}/\textbf{4.3} & \textbf{1.2}/\textbf{2.9} & \textbf{1.2}/\textbf{3.0} & \textbf{1.1}/\textbf{2.7} & \textbf{0.9}/\textbf{2.3} & \textbf{1.3}/\textbf{3.0} \\
\hline
\multicolumn{7}{c}{Concourse Day} \\
\hline
Camera & Pinhole & Fisheye1 & Fisheye2 & Fisheye3 & 360 & Avg \\
\hline
\hline
Ray-DSAC$^*$Net & 8.8/33.8 & 3.2/15.7 & 2.9/13.2 & 2.8/14.9 & 2.6/14.6 & 4.1/18.4 \\
Ray-ACE & 6.9/26.8 & 2.6/12.6 & 2.3/12.1 & 3.4/15.8 & 5.0/14.5 & 4.0/16.4 \\
Ray-GLACE & 5.5/22.1 & 2.5/13.6 & 2.7/13.7 & 2.8/16.6 & 2.7/12.0 & 3.3/15.6 \\
Ray-ACEG  & \textbf{1.4}/\textbf{6.0} & \textbf{0.9}/\textbf{4.4} & \textbf{1.0}/\textbf{4.5} & \textbf{0.8}/\textbf{4.1} & \textbf{0.7}/\textbf{3.5} & \textbf{1.0}/\textbf{4.5} \\
\hline
\multicolumn{7}{c}{Hall Day} \\
\hline
Camera & Pinhole & Fisheye1 & Fisheye2 & Fisheye3 & 360 & Avg \\
\hline
\hline
Ray-DSAC$^*$Net & 9.8/39.6 & 2.2/10.2 & 2.1/9.6 & 2.1/11.2 & 3.0/11.0 & 3.8/16.3 \\
Ray-ACE & 7.5/28.9 & 1.9/9.9 & 1.7/8.6 & 1.9/10.6 & 3.0/9.6 & 3.2/13.5 \\
Ray-GLACE & 4.0/20.3 & 1.5/7.8 & 1.4/7.4 & 1.7/12.6 & 1.7/6.4 & 2.0/10.9 \\
Ray-ACEG   & \textbf{1.5}/\textbf{6.8} & \textbf{1.1}/\textbf{4.4} & \textbf{1.0}/\textbf{4.2} & \textbf{0.9}/\textbf{4.4} & \textbf{0.9}/\textbf{4.1} & \textbf{1.1}/\textbf{4.8} \\
\hline
\multicolumn{7}{c}{Piatrium Day} \\
\hline
Camera & Pinhole & Fisheye1 & Fisheye2 & Fisheye3 & 360 & Avg \\
\hline
\hline
Ray-DSAC$^*$Net & 19.7/32.1 & 6.0/10.1 & 5.3/9.2 & 6.1/12.1 & 5.5/8.8 & 8.5/14.5 \\
Ray-ACE & 20.7/30.6 & 6.9/11.7 & 6.5/11.6 & 8.2/15.2 & 10.2/13.7 & 10.5/16.6 \\
Ray-GLACE & 9.0/18.8 & 3.9/7.4 & 3.9/6.9 & 4.2/11.6 & 3.7/8.1 & 5.0/10.6 \\
Ray-ACEG  & \textbf{4.2}/\textbf{8.5} & \textbf{2.8}/\textbf{5.1} & \textbf{2.6}/\textbf{4.9} & \textbf{2.7}/\textbf{5.2} & \textbf{2.4}/\textbf{5.7} & \textbf{2.9}/\textbf{5.9} \\
\hline
\end{tabular}%
}
\end{table}

%% file: Table/tab7_aachen.tex
\begin{table}[t]
\centering
\caption{\textbf{Localization on Aachen Day.}
ACE and Ray-ACE use $N$ spatially clustered scene-coordinate maps. We report their scene-specific storage and the percentage of queries localized within each pose threshold.}
\label{tab:aachen}
\vspace{0.5mm}
\begingroup
\footnotesize
\renewcommand{\arraystretch}{0.95}
\setlength{\tabcolsep}{3.5pt}
\begin{adjustbox}{max width=\columnwidth}
\begin{tabular}{l|c|c|ccc}
\hline
 & $N$ & Map [MB] & $(0.25\,\mathrm{m},2^\circ)$ & $(0.5\,\mathrm{m},5^\circ)$ & $(5\,\mathrm{m},10^\circ)$ \\
\hline\hline
\multirow{2}{*}{ACE~\cite{brachmann2023ace}}
 & 4  & 16  & 0.0 & 0.5  & 3.8  \\
 & 50 & 205 & 6.9 & 17.2 & 50.0 \\
\hline
GLACE~\cite{wang2024glace}
 & 1 & 27 & 8.6 & 20.8 & 64.0 \\
\hline
\multirow{2}{*}{Ray-ACE}
 & 4  & 19  & 0.4 & 1.5  & 41.0 \\
 & 50 & 213 & \textbf{14.7} & \textbf{35.8} & \textbf{80.1} \\
\hline
Ray-GLACE
 & 1 & 29 & 14.0 & 31.7 & 72.3 \\
\hline
\end{tabular}
\end{adjustbox}
\endgroup
\end{table}

%% file: Table/tab6_viewgen.tex
\begin{table}[t]
\centering
\caption{\textbf{Virtual-view augmentation on 360Loc Concourse-Day.} ACE-G and
Ray-ACEG are trained with augmented mapping views from the rendered cross cameras and evaluated on all five query cameras. Values are median
translation/rotation errors (m/$^\circ$).}
\label{tab:viewgen}
\vspace{1mm}
\resizebox{\columnwidth}{!}{%
\begin{tabular}{c|ccc|cccccc}
\hline
 & \multicolumn{3}{c|}{Augmented views} & \multicolumn{6}{c}{Query camera} \\
 & 360 & Fish. & Pin. & Pinhole & Fisheye1 & Fisheye2 & Fisheye3 & 360 & Avg \\
\hline
\multirow{4}{*}{\rotatebox[origin=c]{90}{ACE-G~\cite{bruns2025aceg}}}
 & \checkmark & \checkmark & \checkmark & 4.7/9.9 & 4.1/11.2 & 4.1/12.2 & 4.5/11.9 & 8.1/33.5 & 5.1/15.7 \\
 & \checkmark & $\times$ & \checkmark & 3.6/8.8 & 4.9/13.7 & 4.9/14.6 & 5.9/18.6 & 8.0/34.1 & 5.5/18.0 \\
 & \checkmark & \checkmark & $\times$ & 9.8/17.2 & 4.6/12.5 & 4.5/12.7 & 4.6/13.0 & 7.7/32.9 & 6.2/17.7 \\
 & \checkmark & $\times$ & $\times$ & 42.9/70.9 & 42.5/76.0 & 40.7/75.5 & 47.0/68.0 & 7.6/31.9 & 36.1/64.5 \\
\hline
\multirow{4}{*}{\rotatebox[origin=c]{90}{Ray-ACEG}}
 & \checkmark & \checkmark & \checkmark & \textbf{1.4}/\textbf{6.0} & \textbf{0.9}/\textbf{4.4} & \textbf{1.0}/\textbf{4.5} & \textbf{0.8}/\textbf{4.1} & \textbf{0.7}/\textbf{3.5} & \textbf{1.0}/\textbf{4.5} \\
 & \checkmark & $\times$ & \checkmark & 1.4/6.8 & 1.0/4.7 & 1.1/5.0 & 1.0/5.2 & 0.9/4.0 & 1.1/5.2 \\
 & \checkmark & \checkmark & $\times$ & 2.8/10.7 & 1.1/5.0 & 1.0/4.5 & 0.9/4.6 & 0.6/3.3 & 1.3/5.6 \\
 & \checkmark & $\times$ & $\times$  & 6.3/16.9 & 2.3/10.1 & 1.9/9.0 & 1.5/8.4 & 1.0/4.1 & 2.6/9.7 \\
\hline
\end{tabular}%
}
\vspace{-0.2cm}
\end{table}

%% file: Table/tab8_cubemap.tex
\begin{table}[t]
\centering
\caption{\textbf{Cubemap-based SCR on 360Loc Concourse-Day.} Standard ACE and ACE-G are trained and evaluated on perspective cubemap faces, whereas their Ray-SCR counterparts operate directly on native calibrated queries. Values are median translation/rotation errors (m/$^\circ$).}
\label{tab:cubemap}
\vspace{1mm}
\resizebox{\columnwidth}{!}{%
\begin{tabular}{l|cccccc}
\hline
 & \multicolumn{6}{c}{Query camera} \\
 & Pinhole & Fisheye1 & Fisheye2 & Fisheye3 & 360 & Avg \\
\hline
Cubemap-ACE~\cite{brachmann2023ace} & 18.8/36.1 & 10.1/18.7 & 8.7/13.1 & 6.8/11.2 & 3.1/4.4 & 9.5/16.7 \\
Ray-ACE & 6.9/26.8 & 2.6/12.6 & 2.3/12.1 & 3.4/15.8 & 5.0/14.5 & 4.0/16.4 \\
\hline
Cubemap-ACEG~\cite{bruns2025aceg} & 14.5/14.4 & 9.6/13.8 & 9.7/13.1 & 10.0/13.1 & 7.2/8.9 & 10.2/12.7 \\
Ray-ACEG & \textbf{1.4}/\textbf{6.0} & \textbf{0.9}/\textbf{4.4} & \textbf{1.0}/\textbf{4.5} & \textbf{0.8}/\textbf{4.1} & \textbf{0.7}/\textbf{3.5} & \textbf{1.0}/\textbf{4.5} \\
\hline
\end{tabular}
}
\end{table}

%% file: Table/tab5_ray_ladder.tex
\begin{table}[t]
\centering
\small
\caption{\textbf{Component analysis on 360Loc Concourse.} Values are camera-averaged
median translation/rotation errors (m/$^\circ$) for Day and Night queries.}
\label{tab:ray_ladder}
\vspace{1mm}
\resizebox{\linewidth}{!}{
\begin{tabular}{lcccc}
\hline
 & w/o Attention & w/o $\mathcal{L}_{\mathrm{ray}}$ & w/o $\mathcal{L}_{\mathrm{sfm}}$ & Full \\
\hline
Day Avg. &  1.32/5.89 & 1.74/9.16 & 1.53/8.39 & \textbf{0.96}/\textbf{4.54} \\
Night Avg. & 1.54/6.95 & 2.37/11.83 & 1.75/9.28 & \textbf{1.28}/\textbf{5.98} \\
\hline
\end{tabular}
\vspace{-0.2cm}
}
\end{table}

%% file: Section/5_Conclusion.tex
\section{Conclusion}
We presented Ray-SCR, which uses calibrated camera rays to make scene coordinate regression explicitly aware of camera geometry. Our method conditions and locally refines visual features using their viewing rays and formulates scene-coordinate learning directly in ray space. Sparse SfM anchors further provide metric 3D supervision while linking observations across views and camera models to consistent scene coordinates. Across several SCR architectures, Ray-SCR substantially improves cross-camera localization, achieving state-of-the-art results on 360Loc and OmniScenes while also improving the corresponding SCR baseline on Aachen. These results establish calibrated ray conditioning as a practical way to extend compact SCR localization beyond perspective cameras.

%% file: Appendix.tex
% Supplementary spacing; prose and font sizes are unchanged.
\setlength{\intextsep}{6pt}
\setlength{\textfloatsep}{8pt}
\setlength{\abovedisplayskip}{4pt}
\setlength{\belowdisplayskip}{4pt}
\setlength{\abovedisplayshortskip}{0pt}
\setlength{\belowdisplayshortskip}{4pt}

\section{SCR Backbone}
\vspace{-3pt}
\label{app:scr_backbones}

Ray-ACE and Ray-GLACE~\cite{wang2024glace} use ACE's~\cite{brachmann2023ace} frozen pretrained encoder and prediction head, an MLP of $1{\times}1$ convolutions with residual blocks. Ray-DSAC$^*$~\cite{brachmann2022dsacstar} instead trains its encoder end-to-end per scene and uses a head of three $1{\times}1$ convolutions with ReLU activations. Ray Adapter uses the same configuration across backbones and differs only in the width of the features it adapts. It contains $0.9$M parameters for $512$-D features and $1.54$M parameters for the $1024$-D features of Ray-ACEG~\cite{bruns2025aceg}. For Ray-GLACE, the R2Former~\cite{zhu2023r2former} global descriptor is computed once per mapping view and concatenated to the adapted local feature before the head. On 360Loc, Ray-ACE, Ray-GLACE, and Ray-DSAC$^*$ are optimized with AdamW at fixed learning rates of $2\mathrm{e}{-3}$, $3\mathrm{e}{-3}$, and $1\mathrm{e}{-3}$, respectively. Ray-ACE and Ray-GLACE each cache 8M feature vectors for training, whereas Ray-DSAC$^*$ uses no feature buffer. All three follow the 70{,}000-iteration training budget and the data preparation and supervision settings described for 360Loc in Sec.~\ref{sec:exp_setup}.

\vspace{-3pt}
\section{Dataset Protocols}
\vspace{-3pt}
\label{supp:data}

\vspace{-2pt}
\subsection{OmniScenes}
\vspace{-2pt}
\label{supp:omniscenes_protocol}

\begin{figure*}[p]
\centering
\includegraphics[width=\textwidth]{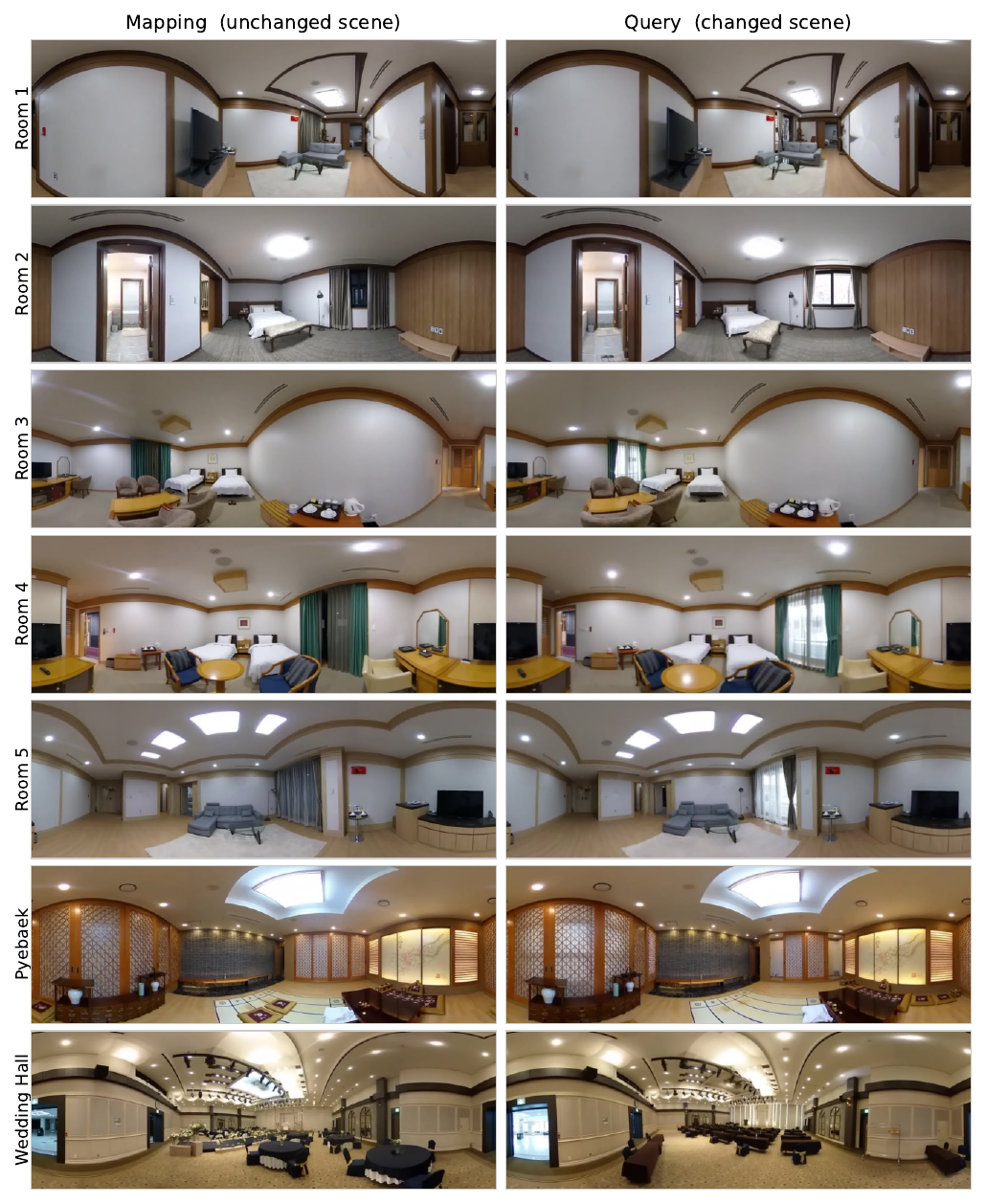}
\caption{\textbf{OmniScenes mapping and query examples.} The mapping panoramas are captured by a robot-mounted camera, while the query panoramas are captured by a handheld camera after rearrangement of the scene.}
\label{fig:omniscenes_pairs}
\end{figure*}

Each of the seven spaces is mapped from its posed robot-mounted panoramas of the unchanged scene (676--923 frames per space) and queried with the handheld panoramas of the Extreme split captured after rearrangement (44--186 frames per space), as illustrated in Fig.~\ref{fig:omniscenes_pairs}. Sparse SfM points are obtained by triangulating SIFT~\cite{lowe2004distinctive} tracks from the mapping frames. Each space is trained for $170$k iterations with Ray-ACE, $110$k with Ray-GLACE, $70$k with Ray-DSAC$^*$, and $40$k with Ray-ACEG, with all other settings following the 360Loc implementation details.

\vspace{-2pt}
\subsection{Aachen}
\vspace{-2pt}
\label{supp:aachen_protocol}

Mapping and query images are resampled using their calibrations to a common focal length of $480$ pixels, standardizing the perspective scale while retaining fields of view. Virtual views are sampled by random rotations within $\pm8^\circ$ about each mapping camera. We use the sparse SfM points provided in the Aachen reference reconstruction as scene-coordinate anchors. Ray-ACE and Ray-GLACE are trained for $30$k and $100$k iterations per map, with buffer sizes of $32$M and $64$M, respectively.

\vspace{-3pt}
\section{Additional Method Details}
\vspace{-3pt}
\label{supp:method}

\vspace{-2pt}
\subsection{Sparse-SfM Anchor on 360Loc}
\vspace{-2pt}
\label{sec:sfm_details}
We construct sparse SfM points by matching SuperPoint~\cite{detone2018superpoint} features with LightGlue~\cite{lindenberger2023lightglue} across perspective views sampled from the mapping panoramas ($95^\circ$ field of view, $1024\times768$ resolution). Matching uses the 16 nearest mapping frames. After retaining the most reliable observations based on reprojection error as described in Sec.~\textcolor{cvprblue}{3.2}, each triangulated point $\mathbf{X}_p$ is assigned a confidence that increases with the angular spread and the number of supporting views. When several points fall on the same feature location, their confidence-weighted average defines the anchor target $\mathbf{X}^{\mathrm{sfm}}_i$. We sum their confidences to obtain $w_i$ in Eq.~\textcolor{cvprblue}{12}, and normalize the weights to unit mean over the valid anchors of each image.

\vspace{-2pt}
\subsection{Pose Estimation}
\vspace{-2pt}
\label{app:pose_estimation}

We provide the explicit angular-error and pose-refinement formulations used in Sec.~\textcolor{cvprblue}{3.5}. For a pose $\mathbf{T}_{w\leftarrow c} =(\mathbf{R}_{w\leftarrow c},\mathbf{t}_w)$, the angular error of the bearing--3D correspondence $(\mathbf{r}_i,\hat{\mathbf{X}}_i)$ is
\begin{equation} e_i(\mathbf{T}_{w\leftarrow c}) = \angle\left( \mathbf{r}_i, \frac{ \mathbf{R}_{w\leftarrow c}^{\top} (\hat{\mathbf{X}}_i-\mathbf{t}_w) }{ \left\| \mathbf{R}_{w\leftarrow c}^{\top} (\hat{\mathbf{X}}_i-\mathbf{t}_w) \right\|_2 } \right). \label{eq:pose_angular_error} \end{equation}

The RANSAC estimate is refined over its inlier correspondences $\mathcal{I}$ as
\begin{equation} \mathbf{T}^{*}_{w\leftarrow c} = \arg\min_{\mathbf{T}_{w\leftarrow c}} \sum_{i\in\mathcal{I}} \rho\!\left( e_i(\mathbf{T}_{w\leftarrow c})^2 \right), \label{eq:pose_refinement} \end{equation}
where $\rho(\cdot)$ is a Huber loss used for nonlinear refinement. Hypotheses are generated with the P3P solver~\cite{kneip2011p3p} inside RANSAC, using an angular inlier threshold of $0.5^\circ$ and $1{,}000$ iterations.

\vspace{-3pt}
\section{Transformer Window Size Ablation}
\vspace{-3pt}
\label{supp:window_size}
Table~\ref{tab:transformer_window} varies the window size $w$ of the local Transformer on Concourse. Accuracy improves as the window grows from $w{=}1$ to $w{=}8$. The main setting $w{=}8$ achieves the lowest Day error, while $w{=}16$ achieves the lowest Night error.

\begin{table}[!ht]
\centering
\small
\captionsetup{skip=2pt}
\caption{\textbf{Transformer window size ablation on 360Loc Concourse.} Values are camera-averaged median translation/rotation errors (m/$^\circ$) for Day and Night queries.}
\label{tab:transformer_window}
\resizebox{\columnwidth}{!}{%
\begin{tabular}{@{}lccccc@{}}
\hline
 & $w=1$ & $w=2$ & $w=4$ & $w=8$ (main) & $w=16$ \\
\hline
Day Avg. & 1.26/5.72 & 1.25/6.00 & 1.11/5.60 & \textbf{0.96}/\textbf{4.54} & 0.98/4.77 \\
Night Avg. & 1.52/6.68 & 1.48/6.82 & 1.39/6.49 & 1.28/5.98 & \textbf{1.20}/\textbf{5.60} \\
\hline
\end{tabular}%
}
\vspace{-0.2cm}
\end{table}

\vspace{\fill}
\section{Limitations}
\vspace{-3pt}
Ray-SCR requires calibrated cameras. Since ray field is computed from the intrinsic calibration, calibration errors bias the mapping and camera pose estimation. Furthermore, the ray-conditioning modules add an inference overhead of $1$\,ms per image for Ray-ACE and $7$\,ms for Ray-ACEG.

\clearpage
